\documentclass[final]{cvpr}

\usepackage{times}
\usepackage{epsfig}
\usepackage{graphicx}
\usepackage{amsmath}
\usepackage{amssymb}
\usepackage{subfigure}
\usepackage{color}
\usepackage{multirow}
\usepackage{diagbox}
\usepackage{array}
\usepackage{booktabs}
\usepackage{overpic}
\usepackage[table]{xcolor}
\makeindex

\DeclareGraphicsExtensions{.pdf,.jpg,.png}

\def\ie{\emph{i.e.,~}}
\def\eg{\emph{e.g.,~}}
\def\etal{{\em et al.~}}

\usepackage{pifont}
\usepackage{overpic}

\newcommand{\myPara}[1]{\vspace{.05in}\noindent\textbf{#1.}}
\usepackage{url}

\definecolor{mygreen}{RGB}{0,100,0}
\definecolor{myred}{RGB}{150,0,0}
\usepackage{hyperref}
\hypersetup{pagebackref=false,colorlinks=true,linkcolor=myred,citecolor=mygreen,bookmarks=false,urlcolor=black}

\def\cvprPaperID{2807} 
\def\confYear{CVPR 2021}

\begin{document}

\title{Coordinate Attention for Efficient Mobile Network Design}

\author{Qibin Hou$^1$ \qquad Daquan Zhou$^1$ \qquad Jiashi Feng$^1$ \\
  $^1$National University of Singapore \\
  \small \url{{andrewhoux, zhoudaquan21}@gmail.com}
}

\maketitle
\thispagestyle{empty}

\begin{abstract}
Recent studies on mobile network design have demonstrated the remarkable effectiveness of channel attention (\eg the Squeeze-and-Excitation attention) for lifting model performance, but they generally neglect the positional information, which is important for generating spatially selective attention maps.
In this paper, we propose a novel attention mechanism for mobile networks by embedding positional information into channel attention, which we call ``coordinate attention''.
Unlike channel attention that transforms a feature tensor to a single feature vector via 2D global pooling, the coordinate attention factorizes channel attention into two 1D feature encoding processes that aggregate features along the two spatial directions, respectively.
In this way, long-range dependencies can be captured along one spatial direction and meanwhile precise positional information can be preserved along the other spatial direction.
The resulting feature maps are then encoded separately into a pair of direction-aware and position-sensitive attention maps that can be complementarily applied to the input feature map to augment the representations of the objects of interest.
Our coordinate attention is simple and can be flexibly plugged into classic mobile networks, such as MobileNetV2,
MobileNeXt, and EfficientNet with nearly no computational overhead.
Extensive experiments demonstrate that our coordinate attention is not only beneficial to ImageNet classification but more interestingly, behaves better in down-stream tasks, such as object detection and semantic segmentation.
Code is available at \url{https://github.com/Andrew-Qibin/CoordAttention}.

\end{abstract}

\section{Introduction} \label{sec:introduction}



Attention mechanisms, used to tell a model ``what'' and ``where'' to attend, have been extensively studied~\cite{xu2015show,mnih2014recurrent} and widely deployed for boosting the performance of modern deep neural networks
\cite{hu2018squeeze,woo2018cbam,cao2019gcnet,liu2020improving,fu2019dual,hou2020strip}.
%
%
However, their application for mobile networks (with limited model size) significantly lags behind that for large networks \cite{simonyan2014very,He2016,xie2017aggregated}.
This is mainly because 
the computational overhead brought by most attention mechanisms is not affordable for mobile networks.

%


%

\begin{figure}[tp]
  \centering
  \includegraphics[width=\linewidth]{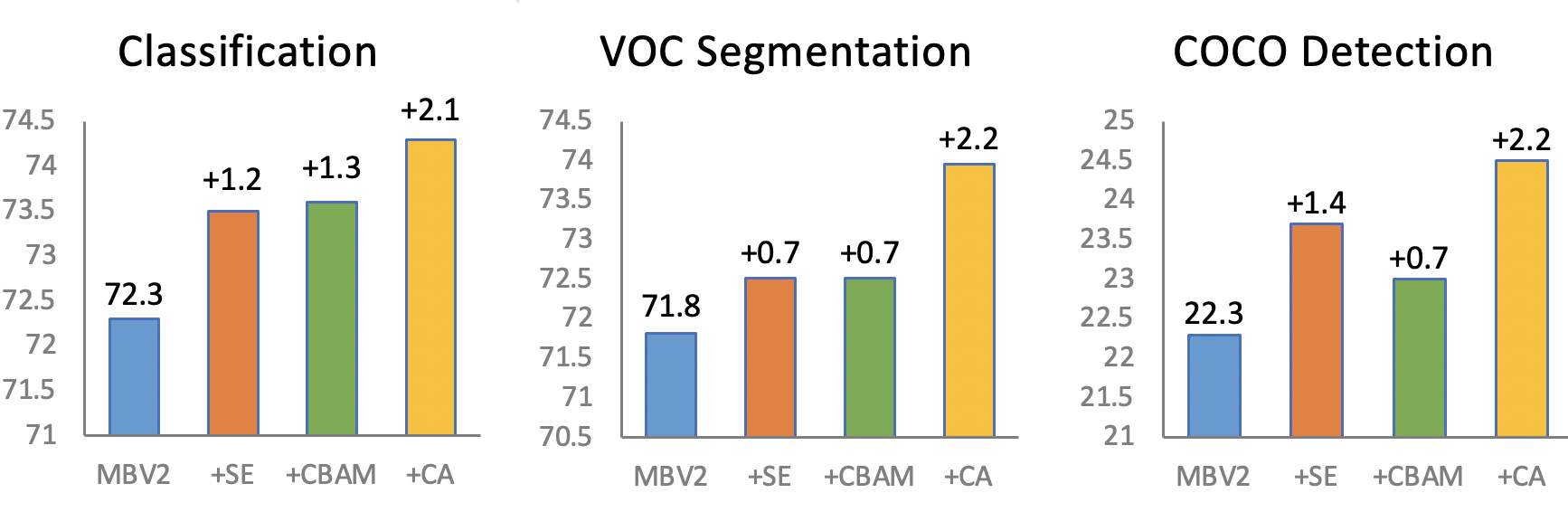}
  \caption{Performance of different attention methods on three classic vision tasks. 
  The y-axis labels from left to right are top-1 accuracy, mean IoU, and AP, respectively.
  Clearly, our approach not only achieves the best result in ImageNet classification \cite{russakovsky2015imagenet} against the SE block \cite{hu2018squeeze} and CBAM \cite{woo2018cbam} 
  but performs even better in down-stream tasks, like semantic segmentation~\cite{everingham2015pascal} and COCO object detection~\cite{lin2014microsoft}.
  Results are based on MobileNetV2 \cite{sandler2018mobilenetv2}.
  }\label{fig:teasor}
\end{figure}

\begin{figure*}[tp]
  \centering
  \includegraphics[width=0.95\linewidth]{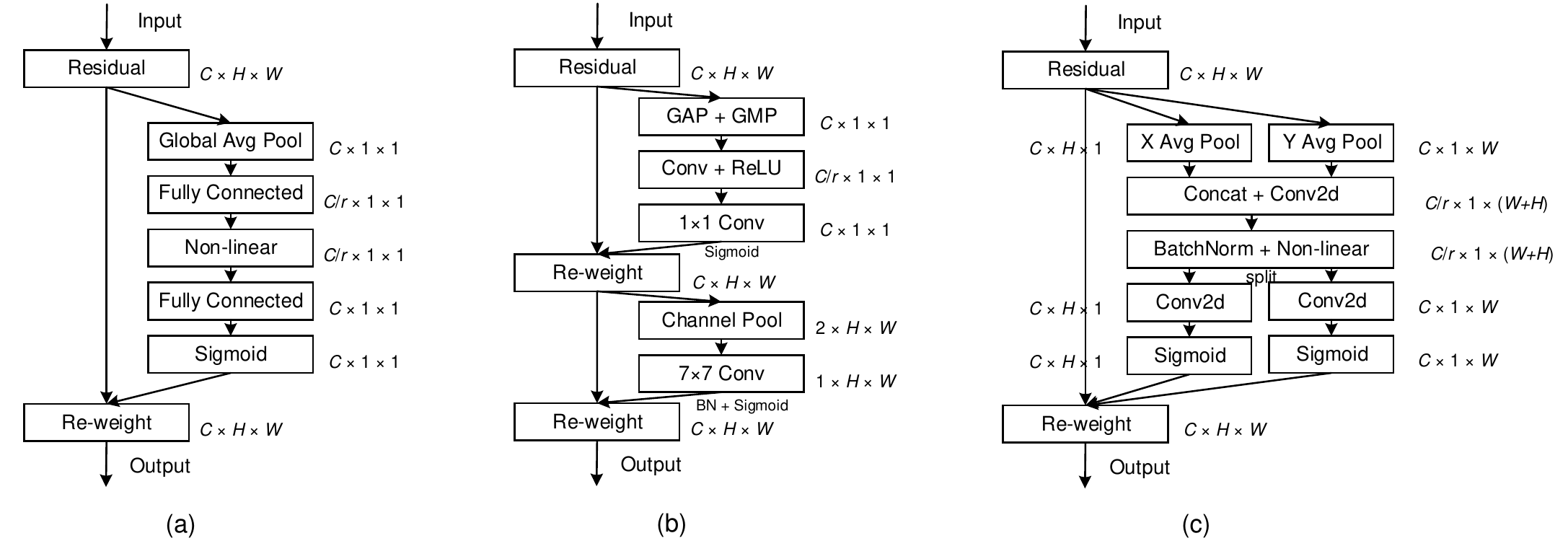}
  \caption{Schematic comparison of the proposed coordinate attention block (c) to the classic SE channel attention block \cite{hu2018squeeze} (a) and CBAM \cite{woo2018cbam} (b). Here, ``GAP''
  and ``GMP'' refer to the global average pooling and global max pooling, respectively. `X Avg Pool'
  and 'Y Avg Pool' refer to 1D horizontal global pooling
  and 1D vertical global pooling, respectively.
  }\label{fig:comp}
\end{figure*}

Considering the restricted computation capacity of mobile networks, to date, the most popular attention mechanism for mobile networks is still the Squeeze-and-Excitation (SE) attention~\cite{hu2018squeeze}.
It computes channel attention with the help of 2D global pooling and provides notable performance gains at considerably low computational cost.
However, the SE attention only considers encoding inter-channel information but neglects the importance of positional information, which is critical to capturing object structures in vision tasks~\cite{wang2020axial}.
Later works, such as BAM \cite{park2018bam} and CBAM \cite{woo2018cbam},
attempt to exploit positional information by reducing the channel dimension of the input tensor and then computing spatial attention using convolutions as shown in Figure~\ref{fig:comp}(b).
However, convolutions can only capture local relations but fail in modeling long-range dependencies that are essential for vision tasks~\cite{zhao2016pyramid,hou2020strip}.

In this paper, beyond the first works, 
we propose a novel and efficient attention mechanism by embedding positional information into channel attention to enable mobile networks to attend over large regions while avoiding incurring significant computation overhead.
%
To alleviate the positional information loss caused by the 2D global pooling, we factorize channel attention into two parallel 1D feature encoding processes to effectively integrate spatial coordinate information into the generated attention maps. 
%
Specifically, our method exploits two 1D global pooling operations to respectively aggregate the input features along the vertical and horizontal directions into two separate direction-aware feature maps.
These two feature maps with embedded direction-specific information are then separately encoded into two attention maps, each of which captures long-range dependencies of the input feature map along one spatial direction.
The positional information can thus be preserved in the generated attention maps.
%
%
Both attention maps are then applied to the input feature map via multiplication to emphasize the representations of interest.
We name the proposed attention method as \emph{coordinate attention} as its operation distinguishes spatial direction (\ie coordinate) and generates coordinate-aware attention maps.
%
%
%


Our coordinate attention offers the following advantages.
First of all, it captures not only cross-channel but also direction-aware and position-sensitive information, which helps models to 
more accurately locate and recognize the objects of interest.
Secondly, our method is flexible and light-weight, and can be easily plugged into classic building blocks of mobile networks, such as the inverted residual block proposed in MobileNetV2~\cite{sandler2018mobilenetv2} and the sandglass block proposed in MobileNeXt~\cite{daquan2020rethinking},
to augment the features by emphasizing informative representations.
Thirdly, as a pretrained model, our coordinate attention can bring  significant performance gains to down-stream tasks with mobile networks, especially for those with dense predictions (\eg semantic segmentation),  which we will show in our experiment section.

To demonstrate the advantages of the proposed approach over previous attention methods for mobile networks,
we conduct extensive experiments in both ImageNet classification~\cite{russakovsky2015imagenet} and popular down-stream tasks, including object detection and semantic segmentation.
With a comparable amount of learnable parameters and computation, our network achieves 0.8\% performance gain in top-1 classification accuracy on ImageNet.
In object detection and semantic segmentation, we also observe significant improvements compared to models with other attention mechanisms as shown in Figure~\ref{fig:teasor}.
%
%
We hope our simple and efficient design could facilitate the development of attention mechanisms for mobile networks in the future.




\section{Related Work} \label{sec:related_work}

In this section, we give a brief literature review of this paper, including prior works on efficient network architecture design and attention or non-local models.

\subsection{Mobile Network Architectures}

%
Recent state-of-the-art mobile networks are mostly based on the depthwise separable convolutions~\cite{howard2017mobilenets} and  the inverted residual block~\cite{sandler2018mobilenetv2}.
%
%
HBONet~\cite{li2019hbonet} introduces down-sampling operations inside each inverted residual block for modeling the representative spatial information.
ShuffleNetV2~\cite{ma2018shufflenet} uses a channel split module
and a channel shuffle module before and after the inverted residual block.
Later, MobileNetV3~\cite{howard2019searching} combines with neural architecture search algorithms~\cite{zoph2018learning} to search for optimal activation functions and the expansion ratio of inverted residual blocks at different depths.
Moreover, MixNet~\cite{tan2019mixconv}, EfficientNet~\cite{tan2019efficientnet} and ProxylessNAS \cite{cai2018proxylessnas} also adopt different searching strategies to search for either the optimal kernel sizes  of the depthwise separable convolutions or scalars to control the network weight in terms of expansion ratio, input resolution, network depth and width.
More recently, Zhou \etal~\cite{daquan2020rethinking} rethought the way of exploiting depthwise separable convolutions and proposed MobileNeXt that adopts a classic bottleneck structure for mobile networks.

\subsection{Attention Mechanisms}

Attention mechanisms \cite{tsotsos1990analyzing,tsotsos2011computational} have been proven helpful in a variety of computer vision tasks, such as image classification~\cite{hu2018squeeze,hu2018gather,woo2018cbam,bello2019attention} and image segmentation~\cite{hou2020strip,huang2018ccnet,fu2019dual}.
One of the successful examples is SENet~\cite{hu2018squeeze}, which simply squeezes each 2D feature map to efficiently build inter-dependencies
among channels.
CBAM~\cite{woo2018cbam} further advances this idea by introducing spatial information encoding via
convolutions with large-size kernels.
Later works, like GENet~\cite{hu2018gather}, GALA~\cite{linsley2018global}, AA~\cite{bello2019attention}, and TA~\cite{misra2021rotate}, extend this idea by adopting different spatial attention mechanisms or designing advanced attention blocks.

Non-local/self-attention networks are recently very popular due to their capability of building spatial or channel-wise attention.
Typical examples include  NLNet~\cite{wang2018non}, GCNet~\cite{cao2019gcnet}, $A^2$Net~\cite{chen20182}, SCNet~\cite{liu2020improving}, GSoP-Net~\cite{gao2019global}, or CCNet~\cite{huang2018ccnet}, all of which exploit non-local mechanisms to capture different types of spatial information.
However, because of the large amount of computation inside the self-attention modules, they are often adopted in large models~\cite{He2016,xie2017aggregated} but not suitable for mobile networks.

Different from these approaches that
leverage expensive and heavy non-local or self-attention blocks, our approach considers a more efficient way of capturing positional information and channel-wise relationships to augment the feature representations for mobile networks.
By factorizing the 2D global pooling operations into two one-dimensional encoding processes, our approach performs much better than other attention methods
with the lightweight property (\eg SENet~\cite{hu2018squeeze}, CBAM~\cite{woo2018cbam}, and TA~\cite{misra2021rotate}).

\section{Coordinate Attention} \label{sec:method}

A \emph{coordinate attention} block can be viewed  as a computational unit that aims to enhance the expressive power of the learned features for mobile networks.
It can take any intermediate feature tensor
$\mathbf{X} = [\mathbf{x}_1, \mathbf{x}_2, \ldots, \mathbf{x}_{C}] \in \mathbb{R}^{C \times H \times W}$ as input
and outputs a transformed tensor with augmented representations
$\mathbf{Y} = [\mathbf{y}_1, \mathbf{y}_2, \ldots, \mathbf{y}_{C}]$ of the same size to $\mathbf{X}$.
%
%
To provide a clear description of the proposed coordinate attention, we first revisit the SE attention, which is widely used in mobile networks.
%

\subsection{Revisit Squeeze-and-Excitation Attention}

As demonstrated in \cite{hu2018squeeze}, the standard convolution itself is difficult to model the channel relationships.
Explicitly building channel inter-dependencies can increase the model sensitivity to the informative channels that contribute more to the final classification decision.
Moreover, using global average pooling can also assist the model in capturing global information, which is a lack for convolutions.

Structurally, the SE block can be decomposed into two steps:
squeeze and excitation, which are designed for global information embedding and adaptive recalibration of channel relationships, respectively.
Given the input $\mathbf{X}$, the squeeze step for the $c$-th channel can be formulated as follows:
\begin{equation} \label{eqn:gap}
    z_c = \frac{1}{H \times W} \sum_{i=1}^{H}\sum_{j=1}^{W} x_c(i, j),
\end{equation}
where $z_c$ is the output associated with the $c$-th channel.
The input $\mathbf{X}$ is directly from a convolutional layer with a fixed kernel size and hence can be viewed as a collection of local descriptors.
The squeeze operation makes collecting global information possible.

The second step, excitation, aims to fully capture channel-wise dependencies, which can be formulated as 
\begin{equation}
    \hat{\mathbf{X}} = \mathbf{X} \cdot \sigma(\hat{\mathbf{z}}),
\end{equation}
where $\cdot$ refers to channel-wise multiplication, $\sigma$ is the sigmoid function, and $\hat{\mathbf{z}}$ is the result generated by a transformation function, which is formulated as follows:
\begin{equation}
    \hat{\mathbf{z}} = T_2(\text{ReLU}(T_1(\mathbf{z}))).
\end{equation}
Here, $T_1$ and $T_2$ are two linear transformations that can be learned to capture the importance of each channel.

The SE block has been widely used in recent mobile networks \cite{hu2018squeeze,chen2018searching,tan2019efficientnet}
and proven to be a key component for
achieving state-of-the-art performance.
However, it only considers reweighing the importance of each channel by modeling channel relationships but neglects positional information, which as we will prove experimentally in Section~\ref{sec:experiments} to be important for generating spatially selective attention maps.
In the following, we introduce a novel attention block, which takes into account both inter-channel
relationships and positional information.

\subsection{Coordinate Attention Blocks}

%

%
Our coordinate attention encodes both channel relationships and long-range dependencies with precise positional information in two steps:
coordinate information embedding and coordinate attention generation.
The diagram of the proposed coordinate attention block can be found in the right part of Figure~\ref{fig:comp}.
In the following, we will describe it in detail.

\subsubsection{Coordinate Information Embedding}


The global pooling is often used in channel attention to encode spatial information globally, but it squeezes global spatial information into a channel descriptor and hence is difficult to preserve positional information, which is essential for capturing spatial structures in vision tasks.
To encourage attention blocks to capture long-range interactions spatially with precise positional information,
we factorize the global pooling as formulated in Eqn.~(\ref{eqn:gap})
into a pair of 1D feature encoding operations. 
Specifically, given the input $\mathbf{X}$, we use two spatial extents of pooling kernels $(H, 1)$ or $(1, W)$ to encode each channel along the horizontal coordinate and the vertical coordinate, respectively.
Thus, the output of the $c$-th channel at height $h$ can be formulated as
\begin{equation} \label{eqn:pool_h}
z_{c}^h(h) = \frac{1}{W} \sum_{0 \le i < W}x_c(h, i).
\end{equation}
Similarly, the output of the $c$-th channel at width $w$ can be written as
\begin{equation} \label{eqn:pool_v}
z_{c}^w(w) = \frac{1}{H} \sum_{0 \le j < H}x_c(j, w).
\end{equation}

The above two transformations aggregate features along the two spatial directions respectively, yielding a pair of direction-aware feature maps.
This is rather different from the squeeze operation (Eqn.~(\ref{eqn:gap})) in channel attention methods
that produce a single feature vector.
These two transformations also allow our attention block to capture long-range dependencies along one spatial direction and preserve precise positional information along the other spatial direction, which helps the networks more accurately locate the objects of interest.
%
%

\subsubsection{Coordinate Attention Generation}

As described above, Eqn.~(\ref{eqn:pool_h}) and Eqn.~(\ref{eqn:pool_v}) enable a global receptive field and encode precise positional information.
To take advantage of the resulting expressive representations, we present the second transformation, termed 
coordinate attention generation.
Our design refers to the following three criteria.
First of all, the new transformation should be as simple and cheap
as possible regarding the applications in mobile environments.
Second, it can make full use of the captured positional information so that the regions of interest can be accurately highlighted.
Last but not the least, it should also be able to effectively capture inter-channel relationships, which has been demonstrated essential in existing studies \cite{hu2018squeeze,woo2018cbam}.
%

Specifically, given the aggregated feature maps produced by Eqn.~\ref{eqn:pool_h} and Eqn.~\ref{eqn:pool_v},
we first concatenate them and then send them to a shared $1\times1$ convolutional transformation function $F_1$,
yielding
\begin{equation} \label{eqn:transformation_1}
\mathbf{f} = \delta(F_1([\mathbf{z}^h, \mathbf{z}^w])),
\end{equation}
where $[\cdot,\cdot]$ denotes the concatenation operation along the spatial dimension, $\delta$ is a non-linear activation function and $\mathbf{f}\in \mathbb{R}^{C/r \times (H + W)}$ is the intermediate feature map that encodes spatial information in both the horizontal direction and the vertical direction.
Here, $r$ is the reduction ratio for controlling the block size as in the SE block.
We then split $\mathbf{f}$ along the spatial dimension into two separate tensors $\mathbf{f}^h\in \mathbb{R}^{C/r \times H}$ and $\mathbf{f}^w\in \mathbb{R}^{C/r \times W}$.
%
Another two $1\times1$ convolutional transformations $F_h$ and $F_w$ are utilized to separately transform
$\mathbf{f}^h$ and $\mathbf{f}^w$ to tensors with the same channel number to the input $\mathbf{X}$, yielding
\begin{align} \label{eqn:transformation_2}
\mathbf{g}^h &= \sigma(F_h(\mathbf{f}^h)), \\
\mathbf{g}^w &= \sigma(F_w(\mathbf{f}^w)).
\end{align}
Recall that $\sigma$ is the sigmoid function.
To reduce the overhead model complexity, we often reduce the channel number of $\mathbf{f}$ with an appropriate reduction ratio $r$ (\eg 32).
We will discuss the impact of different reduction ratios on the performance in our experiment section.
The outputs $\mathbf{g}^h$ and $\mathbf{g}^w$ are then
expanded and used as attention weights, respectively.
Finally, the output of our coordinate attention block $\mathbf{Y}$ can be written as
\begin{equation} \label{eqn:attention}
y_c(i, j) = x_c(i, j) \times g_c^h(i) \times g_c^w(j).
\end{equation}

\myPara{Discussion} Unlike channel attention that only focuses on reweighing the importance of different channels, 
our coordinate attention block also considers encoding the spatial information.
As described above, the attention along both the horizontal and vertical directions is simultaneously applied to the input tensor.
Each element in the two attention maps reflects whether the object of interest exists in the corresponding row and column.
This encoding process allows our coordinate attention to more accurately locate the exact position of the object of interest and hence helps the whole model to recognize better.
We will demonstrate this exhaustively in our experiment section.

\begin{figure}[tp]
  \centering
  \includegraphics[width=\linewidth]{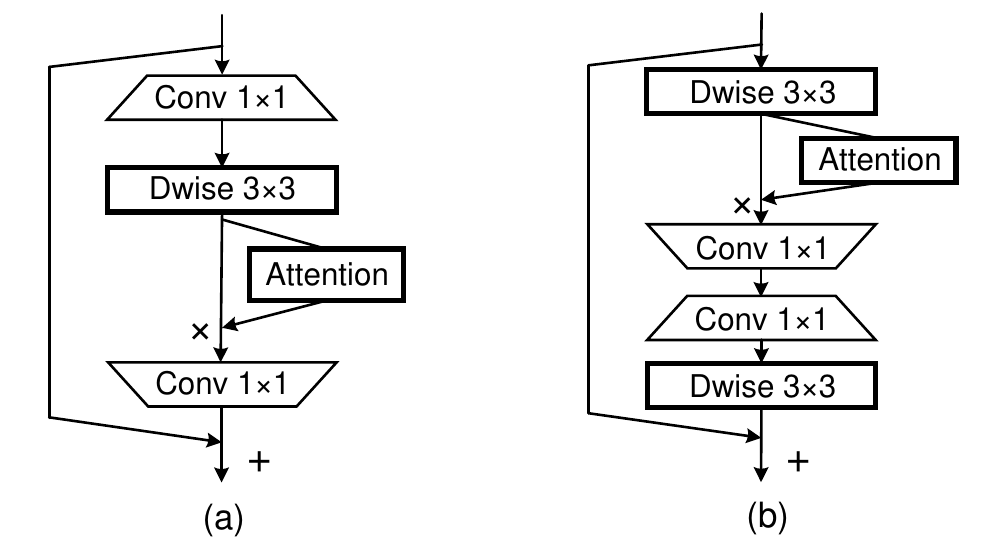}
  \caption{Network implementation for different network architectures. (a) Inverted residual block proposed in MobileNetV2 \cite{sandler2018mobilenetv2};
  (b) Sandglass bottleneck block proposed in MobileNeXt \cite{daquan2020rethinking}.
  }\label{fig:exemplars}
\end{figure}

\subsection{Implementation}

As the goal of this paper is to investigate a better way to augment the convolutional features for mobile networks, here we take two classic light-weight architectures with different types of residual blocks (\ie MobileNetV2 \cite{sandler2018mobilenetv2} and 
MobileNeXt \cite{daquan2020rethinking}) as examples to demonstrate the advantages of the proposed coordinate attention block over other famous light-weight attention blocks.
Figure~\ref{fig:exemplars} shows how we plug attention blocks into the inverted residual block in MobileNetV2 and the sandglass block in MobileNeXt.


\section{Experiments} \label{sec:experiments}
In this section, we first describe our experiment settings and then conduct a series of ablation experiments to demonstrate the contribution of each component in the proposed coordinate attention to the performance.
Next, we compare our approach with some attention based methods.
Finally, we report the results of the proposed approach compared to other attention based methods on object detection and semantic segmentation.
\subsection{Experiment Setup}

We use the PyTorch toolbox \cite{paszke2019pytorch}
to implement all our experiments.
During training, we use the standard SGD optimizer
with decay and momentum of 0.9 to train all the models.
The weight decay is set to $4\times 10^{-5}$ always.
The cosine learning schedule with an initial learning rate of 0.05 is adopted.
We use four NVIDIA GPUs for training and the batch size is set to 256.
Without extra declaration, we take MobileNetV2 as our baseline and train all the models for 200 epochs.
For data augmentation, we use the same methods as
in MobileNetV2.
We report results on the ImageNet dataset \cite{russakovsky2015imagenet} 
in classification.


\begin{table}[t]
    \centering
    \small
    \setlength\tabcolsep{1mm}
    \renewcommand{\arraystretch}{1.1}
    \caption{Result comparisons under different experiment settings of the
    proposed coordinate attention. Here, $r$ is the reduction ratio
    and the baseline result is based on the MobileNetV2 model.
    As can be seen, the model with either the horizontal (X) attention or the vertical (Y) attention added
    achieves the same performance as the one with SE attention. However, when taking both horizontal and vertical
    attentions into account (coordinate attention), our approach yields
    the best result. The latency is tested on a Google Pixel 4 device.}
    \begin{tabular}{lccccccc} \midrule[1pt]
    Settings & Param. & M-Adds & $r$ & Latency & Top-1 (\%)  \\ \midrule[1pt]
    Baseline  &   3.5M      &  300M   & - & 14-16ms &  72.3     \\
    + SE &   3.89M      &  300M & 24  & 16-18ms & $73.5_{+1.2}$     \\
    + X Attention &   3.89M &   300M & 24 & 16-18ms & $73.5_{+1.2}$  \\
    + Y Attention &   3.89M &   300M & 24 & 16-18ms & $73.5_{+1.2}$ \\
    + Coord. Attention &   3.95M & 310M & 32 & 17-19ms & $\mathbf{74.3_{+2.0}}$ \\ \bottomrule[1pt]
    \end{tabular}
    \label{tab:comp_set}
\end{table}

\subsection{Ablation Studies}

\myPara{Importance of coordinate attention}
To demonstrate the performance of the proposed coordinate
attention, we perform a series of ablation experiments,
the corresponding results of which are all listed 
in Table~\ref{tab:comp_set}.
We remove either the horizontal attention
or the vertical attention from the coordinate attention
to see the importance of encoding coordinate information.
As shown in Table~\ref{tab:comp_set}, the model with attention
along either direction has comparable performance to
the one with the SE attention.
However, when both the horizontal attention
and the vertical attention are incorporated,
we obtain the best result as highlighted in
Table~\ref{tab:comp_set}.
These experiments reflect that with comparable
learnable parameters and computational cost,
coordinate information embedding is more helpful
for image classification.

\begin{table}[t]
    \centering
    \small
    \setlength\tabcolsep{1mm}
    \renewcommand{\arraystretch}{1.1}
    \caption{Comparisons of different attention methods under different weight multipliers when taking MobileNetV2 as the baseline. }
    \begin{tabular}{lccccccc} \midrule[1pt]
    Settings & Param. (M) & M-Adds (M) & Top-1 Acc (\%)  \\ \midrule[1pt]
    MobileNetV2-1.0  &   3.5      &  300    &  72.3     \\
    + SE &   3.89      &  300    &  $73.5_{+1.2}$     \\
    + CBAM &   3.89      &  300    &  $73.6_{+1.3}$     \\
    + CA &   3.95 &   310 & $\mathbf{74.3_{+2.0}}$ \\ \midrule[1pt]
    MobileNetV2-0.75  &   2.5      &  200    &  69.9     \\
    + SE &   2.86      &  210    &  $71.5_{+1.6}$     \\
    + CBAM &   2.86      &  210    &  $71.5_{+1.6}$     \\
    + CA &   2.89 &   210 & $\mathbf{72.1_{+2.2}}$ \\ \midrule[1pt]
    MobileNetV2-0.5  &   2.0      &  100    &  65.4     \\
    + SE &   2.1      &  100    &  $66.4_{+1.0}$     \\
    + CBAM &   2.1      &  100    &  $66.4_{+1.0}$     \\
    + CA &   2.1 &   100 & $\mathbf{67.0_{+1.6}}$ \\ \bottomrule[1pt]
    \end{tabular}
    
    \label{tab:mbv2_weight_mult}
\end{table}

\myPara{Different weight multipliers}
Here, we take two classic mobile networks (including
MobileNetV2 \cite{sandler2018mobilenetv2} with inverted
residual blocks and
MobileNeXt \cite{daquan2020rethinking} with sandglass bottleneck block) as baselines
to see the performance of the proposed approach
compared to the SE attention \cite{hu2018squeeze} and
CBAM \cite{woo2018cbam} under different weight
multipliers.
In this experiment, we adopt three typical
weight multipliers, including $\{1.0, 0.75, 0.5\}$.
As shown in Table~\ref{tab:mbv2_weight_mult},
when taking the MobileNetV2 network as baseline,
models with CBAM have similar results to those 
with the SE attention.
However, models with the proposed coordinate attention
yield the best results under each setting.
Similar phenomenon can also be observed when the
MobileNeXt network is used as listed
in Table~\ref{tab:mnext_weight_mult}.
This indicates that no matter which of the sandglass
bottleneck block or the inverted residual block 
is considered and no matter which weight multiplier
is selected, our coordinate attention performs
the best because of the advanced way to encode positional
and inter-channel information simultaneously.

\begin{table}[t]
    \centering
    \small
    \setlength\tabcolsep{1mm}
    \renewcommand{\arraystretch}{1.1}
    \caption{Comparisons of different attention methods under different weight multipliers when taking MobileNeXt \cite{daquan2020rethinking} as the baseline.}
    \begin{tabular}{lccccccc} \midrule[1pt]
    Settings & Param. (M) & M-Adds (M) & Top-1 Acc (\%)  \\ \midrule[1pt]
    MobileNeXt  &   3.5      &  300    &  74.0     \\
    + SE &   3.89      &  300    &  $74.7_{+0.7}$     \\
    + CA &   4.09 &   330 & $\mathbf{75.2_{+1.2}}$ \\ \midrule[1pt]
    MobileNeXt-0.75  &   2.5      &  210    &  72.0     \\
    + SE &   2.9      &  210    &  $72.6_{+0.6}$     \\
    + CA &   3.0 &   220 & $\mathbf{73.2_{+1.2}}$ \\ \midrule[1pt]
    MobileNeXt-0.5  &   2.1      &  110    &  67.7     \\
    + SE &   2.4      &  110    &  $68.7_{+1.0}$     \\
    + CA &   2.4 &   110 & $\mathbf{69.4_{+1.7}}$ \\ \bottomrule[1pt]
    \end{tabular}
    
    \label{tab:mnext_weight_mult}
\end{table}

\myPara{The impact of reduction ratio $r$}
To investigate the impact of different reduction ratios of
attention blocks on the model performance, we attempt to 
decrease the size of the reduction ratio and see the performance
change.
As shown in Table~\ref{tab:reduction_ratio}, when we reduce
$r$ to half of the original size, the model size increases
but better performance can be yielded.
This demonstrates that adding more parameters by reducing the
reduction ratio matters for improving the model performance.
More importantly, our coordinate attention still performs better
than the SE attention and CBAM in this experiment, reflecting the
robustness of the proposed coordinate attention to the reduction ratio.

\begin{table}[t]
    \centering
    \small
    \setlength\tabcolsep{2.5mm}
    \renewcommand{\arraystretch}{1.1}
    \caption{Comparisons of models equipped with different
    attention blocks under different reduction ratios $r$. 
    The baseline result is based on the MobileNetV2 model.
    Obviously, when the reduction ratio decreases, 
    our approach still yields the best results.}
    \begin{tabular}{lcccc} \midrule[1pt]
    Settings & Param. & M-Adds & $r$ & Top-1 Acc (\%)  \\ \midrule[1pt]
    Baseline  &   3.5M      &  300M   & - &  72.3     \\
    + SE &   3.89M      &  300M & 24  &  $73.5_{+1.2}$     \\
    + CBAM &   3.89M      &  300M & 24  & $73.6_{+1.3}$     \\
    + CA (Ours) &   3.95M & 310M & 32 & $\mathbf{74.3_{+2.0}}$ \\ \midrule[1pt]
    + SE &   4.28M      &  300M & 12  &  $74.1_{+1.8}$     \\
    + CBAM &   4.28M      &  300M & 12  & $74.1_{+1.8}$     \\
    + CA (Ours) &   4.37M & 310M & 16 & $\mathbf{74.7_{+2.4}}$ \\
    \bottomrule[1pt]
    \end{tabular}
    \label{tab:reduction_ratio}
\end{table}


\newcommand{\addFig}[1]{\includegraphics[width=0.185\linewidth]{figures/gradcam/#1}}
\newcommand{\addFigs}[3]{\addFig{ILSVRC2012_val_#1_#2_#3.JPEG}}
\newcommand{\addFigsX}[3]{\addFig{#1-#2-layer4-resnext50-#1-#3.png}}
\newcommand{\addFigsS}[3]{\addFig{#1-#2-layer4-seresnet50-#1-#3.png}}
\newcommand{\addFigsP}[3]{\addFig{#1-#2-layer4-resnet50sp-#1-#3.png}}

\renewcommand{\addFig}[1]{\includegraphics[width=0.185\linewidth]{figures/vis_comp/#1}}
\renewcommand{\addFigs}[3]{\addFig{ILSVRC2012_val_#1_#2_#3.JPEG}}
\begin{figure}
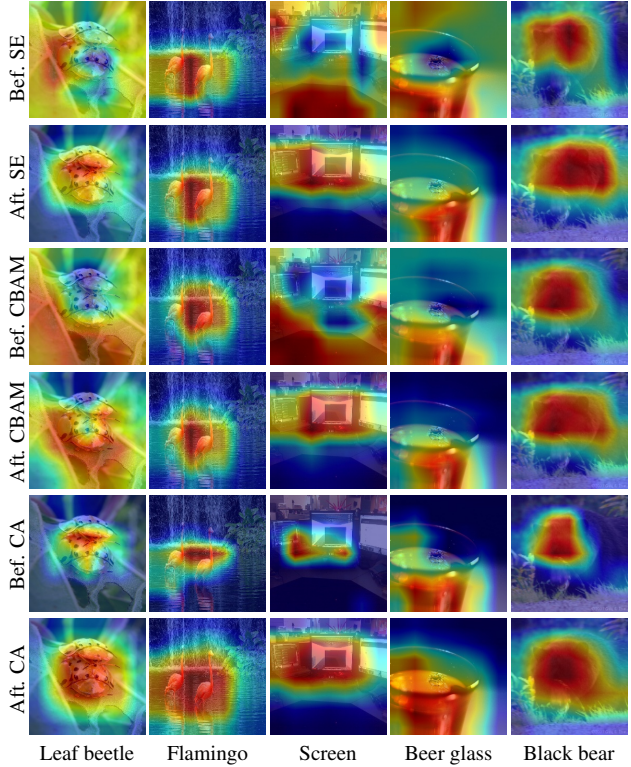

  \centering
  \footnotesize
  \setlength\tabcolsep{0.3mm}
  \renewcommand\arraystretch{1.0}
  \begin{tabular}{cccccc}
     
    \rotatebox{90}{~~~Bef. SE} & 
    \addFigs{00004716}{leaf_beetle}{se_mbv2_5} & 
    \addFigs{00007128}{flamingo}{se_mbv2_5} & 
    \addFigs{00013135}{screen}{se_mbv2_5} & 
    \addFigs{00018397}{beer_glass}{se_mbv2_5} &
    \addFigs{00035591}{American_black_bear}{se_mbv2_5} \\
    \rotatebox{90}{~~~~Aft. SE} &
    \addFigs{00004716}{leaf_beetle}{se_mbv2_6} & 
    \addFigs{00007128}{flamingo}{se_mbv2_6} & 
    \addFigs{00013135}{screen}{se_mbv2_6} & 
    \addFigs{00018397}{beer_glass}{se_mbv2_6} &
    \addFigs{00035591}{American_black_bear}{se_mbv2_6} \\
    
    \rotatebox{90}{~Bef. CBAM} & 
    \addFigs{00004716}{leaf_beetle}{cbam_5} & 
    \addFigs{00007128}{flamingo}{cbam_5} & 
    \addFigs{00013135}{screen}{cbam_5} & 
    \addFigs{00018397}{beer_glass}{cbam_5} &
    \addFigs{00035591}{American_black_bear}{cbam_5} \\
    \rotatebox{90}{~Aft. CBAM} &
    \addFigs{00004716}{leaf_beetle}{cbam_6} & 
    \addFigs{00007128}{flamingo}{cbam_6} & 
    \addFigs{00013135}{screen}{cbam_6} & 
    \addFigs{00018397}{beer_glass}{cbam_6} &
    \addFigs{00035591}{American_black_bear}{cbam_6} \\
    
    \rotatebox{90}{~~~~Bef. CA} & 
    \addFigs{00004716}{leaf_beetle}{canet_5} & 
    \addFigs{00007128}{flamingo}{canet_5} & 
    \addFigs{00013135}{screen}{canet_5} & 
    \addFigs{00018397}{beer_glass}{canet_5} &
    \addFigs{00035591}{American_black_bear}{canet_5} \\
    \rotatebox{90}{~~~~Aft. CA} &
    \addFigs{00004716}{leaf_beetle}{canet_6} & 
    \addFigs{00007128}{flamingo}{canet_6} & 
    \addFigs{00013135}{screen}{canet_6} & 
    \addFigs{00018397}{beer_glass}{canet_6} &
    \addFigs{00035591}{American_black_bear}{canet_6} \\
    
    & Leaf beetle & Flamingo & Screen & Beer glass & Black bear \\
  \end{tabular}
  \vspace{3pt}
  \caption{Visualization of feature maps 
  produced by models with different attention
  methods in the last building block.
  We use Grad-CAM \cite{selvaraju2017grad}
  as our visualization tool.
  Both feature maps before and after each
  attention block are visualized.
  It is obvious that our coordinate attention (CA)
  can more precisely locate the objects of
  interest than other attention methods.}
  \label{fig:vis_comps}
\end{figure}

\subsection{Comparison with Other Methods}

\myPara{Attention for Mobile Networks}
We compare our coordinate attention
with other light-weight attention methods
for mobile networks, including
the widely adopted SE attention \cite{hu2018squeeze} and CBAM \cite{woo2018cbam}
in Table~\ref{tab:mbv2_weight_mult}.
As can be seen, adding the SE attention has
already raised the classification performance
by more than 1\%.
For CBAM, it seems that its spatial attention module
shown in Figure~\ref{fig:comp}(b)
does not contribute in mobile networks compared to 
the SE attention.
However, when the proposed coordinate
attention is considered, we achieve the best results.
We also visualize the feature maps produced
by models with different attention methods
in Figure~\ref{fig:vis_comps}.
Obviously, our coordinate attention can
help better in locating the objects of interest
than the SE attention and CBAM.

We argue that the advantages of the proposed
positional information encoding manner
over CBAM are two-fold.
First, the spatial attention module in CBAM
squeezes the channel dimension to 1, leading
to information loss.
However, our coordinate attention uses
an appropriate reduction ratio to 
reduce the channel dimension in the bottleneck,
avoiding too much information loss.
Second, CBAM utilizes a convolutional layer
with kernel size $7\times7$ to
encode local spatial information while our
coordinate attention encodes global information
by using two complementary 1D global pooling
operations.
This enables our coordinate attention to capture
long-range dependencies among spatial locations
that are essential for vision tasks.
%

\myPara{Stronger Baseline}
To further demonstrate the advantages of the proposed coordinate
attention over the SE attention in more powerful mobile networks,
we take EfficientNet-b0 \cite{tan2019efficientnet} 
as our baseline here.
EfficientNet is based on architecture search algorithms.
and contains SE attention.
To investigate the performance of the proposed coordinate attention
on EfficientNet, we simply replace the SE attention
with our proposed coordinate attention.
For other settings, we follow the original paper.
The results have been listed in Table~\ref{tab:efficientnet}.
Compared to the original EfficientNet-b0 with SE attention included and other methods that have comparable parameters and computations to EfficientNet-b0,
our network with coordinate attention achieves the best result.
This demonstrates that the proposed coordinate attention
can still performance well in powerful mobile networks.

\begin{table}[t]
    \centering
    \small
    \setlength\tabcolsep{1mm}
    \renewcommand{\arraystretch}{1.1}
    \caption{Experimental results when taking the powerful
    EfficientNet-b0 \cite{tan2019efficientnet} as baseline.
    We also compare with other methods that
    have comparable parameters and computations
    to EfficientNet-b0.}
    \begin{tabular}{lcccc} \midrule[1pt]
    Settings & Param. & M-Adds & Top-1 Acc (\%)  \\ \midrule[1pt]
    PNAS \cite{liu2018progressive} & 5.1M & 588M & 72.7 \\
    DARTS \cite{liu2018darts} & 4.7M & 574M & 73.3 \\
    ProxylessNAS-M \cite{cai2018proxylessnas} & 4.1M & 330M & 74.4 \\
    AmoebaNet-A \cite{real2019regularized} & 5.1M & 555M & 74.5 \\
    FBNet-C \cite{wu2019fbnet}  & 5.5M & 375M & 74.9 \\
    MobileNeXt \cite{daquan2020rethinking} & 6.1M & 590M & 76.1 \\
    MNasNet-A3 \cite{tan2019mnasnet} & 5.2M & 403M & 76.7 \\ \midrule[1pt]
    EfficientNet-b0 (w/ SE) \cite{tan2019efficientnet} & 5.3M & 390M & 76.3     \\
    EfficientNet-b0 (w/ CA) &  5.4M & 400M & \textbf{76.9} \\
    \bottomrule[1pt]
    \end{tabular}
    \label{tab:efficientnet}
\end{table}

\subsection{Applications}

In this subsection, we conduct experiments
on both the object detection task and
the semantic segmentation task to explore the transferable capability of the proposed 
coordinate attention against other attention methods.

\begin{table*}[t]
    \centering
    \small
    \setlength\tabcolsep{2.5mm}
    \renewcommand{\arraystretch}{1.2}
    \caption{Object detection results on the COCO validation set. 
    In all experiments here, we use the SSDLite320 detector. As can be seen, the backbone
    model with our coordinate attention 
    achieves the best results in terms of all kinds of measuring metrics.
    Note that all the results are based on
    single-model test.
    Besides hand-designed mobile networks, we
    also show results produced by architecture search-based methods (\ie MobileNetV3 \cite{howard2019searching} and MnasNet-A1 \cite{tan2019mnasnet}).}
    \begin{tabular}{ccccccccccc} \toprule[1pt] 
        No. & Method & Backbone & Param. (M) & M-Adds (B) & AP & AP$_{50}$ & AP$_{75}$ & AP$_{S}$ & AP$_{M}$ & AP$_{L}$ \\ \midrule[1pt]
        1 & SSDLite320  & MobileNetV1 \cite{howard2017mobilenets} & 5.1 & 1.3 & 22.2 & - & - & - & - & - \\
        2 & SSDLite320  & MobileNetV2 \cite{sandler2018mobilenetv2} & 4.3 & 0.8 & 22.3 & 37.4 & 22.7 & 2.8 & 21.2 & 42.8 \\
        3 & SSDLite320  & MobileNetV3 \cite{howard2019searching} & 5.0 & 0.62 & 22.0 & - & - & - & - & - \\
        4 & SSDLite320 & MnasNet-A1 \cite{tan2019mnasnet} & 4.9 & 0.8 & 23.0 & - & - &  3.8 & 21.7 &  42.0 \\
        5 & SSDLite320  & MobileNeXt \cite{daquan2020rethinking} & 4.4 & 0.8 & 23.3 & 38.9 & 23.7 & 2.8 & 22.7 & 45.0 \\ \midrule[1pt]
        6 & SSDLite320  & MobileNetV2 + SE & 4.7 & 0.8 & 23.7 & 40.0 & 24.3 & 2.2 & 25.4 & 44.7 \\ 
        7 & SSDLite320  & MobileNetV2 + CBAM & 4.7 & 0.8 & 23.0 & 38.6 & 23.3 & 2.7 & 22.2 & 44.5 \\ 
        8 & SSDLite320  & MobileNetV2 + CA & 4.8 & 0.8 & \textbf{24.5} & 40.7 & 25.4 & 2.3 & 26.2 & 45.9 \\ 
        \bottomrule[1.0pt]
        \end{tabular}
        \label{tab:coco_detection}
\end{table*}

\begin{table}[t]
    \centering
    \small
    \setlength\tabcolsep{1.2mm}
    \renewcommand{\arraystretch}{1.1}
    \caption{Object detection results on the Pascal VOC 2007 test set. 
    We can observe that when the same SSDLite320 detector is adopted, MobileNetV2 network
    with our coordinate attention added
    achieves better results in terms of mAP.}
    \begin{tabular}{ccccccccc} \toprule[1pt] 
        Backbone &  Param. (M) & M-Adds (B) & mAP (\%) \\ \midrule[1pt]
        MobileNetV2 \cite{sandler2018mobilenetv2}  & 4.3 & 0.8  & 71.7 \\
        MobileNetV2 + SE  & 4.7 & 0.8  & 71.7 \\
        MobileNetV2 + CBAM  & 4.7 & 0.8  & 71.7 \\
        MobileNetV2 + CA  & 4.8 & 0.8  & \textbf{73.1} \\
        \bottomrule[1.0pt]
        \end{tabular}
        \label{tab:voc_detection}
\end{table}

\subsubsection{Object Detection}


\myPara{Implementation Details}
Our code is based on PyTorch and SSDLite \cite{sandler2018mobilenetv2,liu2016ssd}.
%
Following \cite{sandler2018mobilenetv2}, we
connect the first and second layers of SSDLite 
to the last pointwise convolutions with output stride of 16 and 32, respectively
and add the rest SSDLite layers on top of the 
last convolutional layer.
When training on COCO, we set the batch size to 256 and use the synchronized batch normalization.
The cosine learning schedule is used with an initial learning rate of 0.01.
We train the models for totally 1,600,000 iterations.
When training on Pascal VOC, the batch size is
set to 24 and all the models
are trained for 240,000 iterations.
The weight decay is set to 0.9.
The initial learning rate is 0.001, 
which is then divided by 10 at 160,000 and again at 200,000 iterations.
For other settings, readers can refer to
\cite{sandler2018mobilenetv2,liu2016ssd}.

\myPara{Results on COCO}
In this experiment, we follow most previous work
and report results in terms of AP, AP$_{50}$, AP$_{75}$,
AP$_{S}$, AP$_{M}$, and AP$_{L}$, respectively.
In Table~\ref{tab:coco_detection}, we show the results 
produced by different network settings 
on the COCO 2017 validation set.
It is obvious that adding coordinate attention
into MobileNetV2 substantially improve the detection 
results (24.5 v.s. 22.3) with only 0.5M parameters
overhead and nearly the same computational cost.
Compared to other light-weight attention methods,
such as the SE attention and CBAM, 
our version of SSDLite320 achieves the best results 
in all metrics with nearly the same number of
parameters and computations.

Moreover, we also show results produced by
previous state-of-the-art models based on SSDLite320 as listed in
Table~\ref{tab:coco_detection}.
Note that some methods (\eg MobileNetV3 \cite{howard2019searching} and MnasNet-A1 \cite{tan2019mnasnet})
are based on neural architecture search methods
but our model does not.
Obviously, our detection model achieves the best
results in terms of AP compared to other approaches with close parameters and computations.
%

\myPara{Results on Pascal VOC}
In Table~\ref{tab:voc_detection}, we show the 
detection results on Pascal VOC 2007 test set
when different attention methods are adopted.
We observe that the SE attention and CBAM
cannot improve the baseline results.
However, adding the proposed coordinate
attention can largely raise the mean AP
from 71.7 to 73.1.
Both detection experiments on COCO and
Pascal VOC datasets demonstrate that 
classification models with the proposed 
coordinate attention have better
transferable capability compared to 
those with other attention methods.

\begin{table}[t]
    \centering
    \small
    \setlength\tabcolsep{1.8mm}
    \renewcommand{\arraystretch}{1.1}
     \caption{Semantic segmentation results on the Pascal VOC 2012 validation set.
     All the results are based on single-model test and no post-processing tools are used. We can see that the models equipped with
     all attention methods improve
     the segmentation results.
     However, when the proposed coordinate
     attention is used, we achieve the best
     result, which is much better than
     models with other attention methods.
     `Stride' here denotes the output stride
     of the segmentation network.}
    \begin{tabular}{lccc} \toprule[1pt] 
        Backbone & Param. (M) & Stride & mIoU (\%)  \\ \midrule[1pt]
        MobileNetV2 \cite{sandler2018mobilenetv2} & 4.5 & 16 & 70.84 \\
        MobileNetV2 + SE & 4.9 & 16 & 71.69 \\
        MobileNetV2 + CBAM & 4.9 & 16 & 71.28 \\
        MobileNetV2 + CA (ours) & 5.0 & 16 & \textbf{73.32} \\\midrule[1pt]
        MobileNetV2 \cite{sandler2018mobilenetv2} & 4.5 & 8 & 71.82 \\
        MobileNetV2 + SE & 4.9 & 8 & 72.52 \\
        MobileNetV2 + CBAM & 4.9 & 8 & 71.67 \\
        MobileNetV2 + CA (ours) & 5.0 & 8 & \textbf{73.96} \\
        \bottomrule[1pt]
        \end{tabular}
        \label{tab:seg_voc}
\end{table} 

\begin{table}[t]
    \centering
    \small
    \setlength\tabcolsep{1.1mm}
    \renewcommand{\arraystretch}{1.1}
     \caption{Semantic segmentation results on the Cityscapes \cite{cordts2016cityscapes} validation set. 
     We report results on single-model test and
     full image size (\ie $1024 \times 2048$)
     is used for testing.
     We do not use any post-processing tools.}
    \begin{tabular}{lccc} \toprule[1pt] 
        Backbone & Param. (M) & Output Stride & mIoU (\%)  \\ \midrule[1pt]
        MobileNetV2 & 4.5  & 8  & 71.4 \\
        MobileNetV2 + SE & 4.9  & 8  & 72.2 \\
        MobileNetV2 + CBAM & 4.9  & 8  & 71.4 \\
        MobileNetV2 + CA  & 5.0  & 8  & \textbf{74.0}  \\ 
        \bottomrule[1pt]
        \end{tabular}
        \label{tab:seg_city}
\end{table} 

\subsubsection{Semantic Segmentation}

We also conduct experiments on semantic segmentation.
Following MobileNetV2 \cite{sandler2018mobilenetv2},
we utilize the classic DeepLabV3 \cite{chen2017rethinking} 
as an example and compare the proposed approach with
other models to demonstrate the transferable capability of
the proposed coordinate attention in semantic segmentation.
Specifically, we discard the last linear operator and
connect the ASPP to the last convolutional operator.
We replace the standard $3\times3$ convolutional
operators with the depthwise separable convolutions
in the ASPP to reduce the model size considering mobile applications.
The output channels for each branch in ASPP are set to
256 and other components in the ASPP are kept unchanged
(including the $1\times1$ convolution branch and 
the image-level feature encoding branch).
We report results on two widely used semantic segmentation
benchmarks, including Pascal VOC 2012 \cite{everingham2015pascal}
and Cityscapes \cite{cordts2016cityscapes}.
For experiment settings, we strictly follow the DeeplabV3 
paper except for the weight decay that is set to 4e-5.
When the output stride is set to 16, the dilation rates
in the ASPP are \{6, 12, 18\} while \{12, 24, 36\}
when the output stride is set to 8.

\myPara{Results on Pascal VOC 2012}
The Pascal VOC 2012 segmentation benchmark has 
totally 21 classes including one background class.
As suggested by the original paper, we use the split 
with 1,464 images for training and the split with
1,449 images for validation.
Also, as done in most previous work
\cite{chen2017rethinking,chen2014semantic}, 
we augment the training set by adding extra images
from \cite{hariharan2011semantic},
resulting in totally 10,582 images for training.

We show the segmentation results when taking different models
as backbones in Table~\ref{tab:seg_voc}.
We report results under two different output strides, 
\ie 16 and 8.
Note that all the results reported here are not based on
COCO pretraining.
According to Table~\ref{tab:seg_voc}, models equipped with
our coordinate attention performs much better than the vanilla
MobileNetV2 and other attention methods.                 

\myPara{Results on Cityscapes} 
Cityscapes \cite{cordts2016cityscapes} is one of 
the most popular urban street scene segmentation
datasets, containing totally 19
different categories.
Following the official suggestion, we use
the split with 2,975 images for training and 500 images for validation. 
Only the fine-annotated images are used 
for training.
In training, we randomly crop the original
images to $768 \times 768$.
During testing, all images are kept 
the original size ($1024\times2048$).
%

In Table~\ref{tab:seg_city}, we show the segmentation
results produced by models with different 
attention methods on the Cityscapes dataset.
Compared to the vanilla MobileNetV2 and
other attention methods, our coordinate attention can improve the segmentation results
by a large margin with comparable number of learnable parameters.

\myPara{Discussion}
We observe that our coordinate attention
yields larger improvement on semantic
segmentation than ImageNet classification 
and object detection.
We argue that this is because our coordinate
attention is able to capture long-range
dependencies with precise postional information,
which is more beneficial to vision tasks with
dense predictions, such as semantic segmentation.

\section{Conclusions}

In this paper, we present a novel light-weight
attention mechanism for mobile networks, named
coordinate attention.
Our coordinate attention inherits the
advantage of channel attention methods
(\eg the Squeeze-and-Excitation attention)
that model inter-channel relationships
and meanwhile captures long-range dependencies
with precise positional information.
Experiments in ImageNet classification, object
detection and semantic segmentation demonstrate
the effectiveness of our coordination
attention.

\myPara{Acknowledgement}
This research was partially supported by AISG-100E-2019-035, MOE2017-T2-2-151, NUS\_ECRA\_FY17\_P08 and CRP20-2017-0006.

{\small
\bibliographystyle{ieee_fullname}
\bibliography{egbib}

\begin{thebibliography}{10}\itemsep=-1pt

\bibitem{bello2019attention}
Irwan Bello, Barret Zoph, Ashish Vaswani, Jonathon Shlens, and Quoc~V Le.
\newblock Attention augmented convolutional networks.
\newblock {\em arXiv preprint arXiv:1904.09925}, 2019.

\bibitem{cai2018proxylessnas}
Han Cai, Ligeng Zhu, and Song Han.
\newblock Proxylessnas: Direct neural architecture search on target task and
  hardware.
\newblock {\em arXiv preprint arXiv:1812.00332}, 2018.

\bibitem{cao2019gcnet}
Yue Cao, Jiarui Xu, Stephen Lin, Fangyun Wei, and Han Hu.
\newblock Gcnet: Non-local networks meet squeeze-excitation networks and
  beyond.
\newblock In {\em Proceedings of the IEEE International Conference on Computer
  Vision Workshops}, pages 0--0, 2019.

\bibitem{chen2018searching}
Liang-Chieh Chen, Maxwell Collins, Yukun Zhu, George Papandreou, Barret Zoph,
  Florian Schroff, Hartwig Adam, and Jon Shlens.
\newblock Searching for efficient multi-scale architectures for dense image
  prediction.
\newblock In {\em {NeurIPS}}, pages 8699--8710, 2018.

\bibitem{chen2014semantic}
Liang-Chieh Chen, George Papandreou, Iasonas Kokkinos, Kevin Murphy, and Alan~L
  Yuille.
\newblock Semantic image segmentation with deep convolutional nets and fully
  connected crfs.
\newblock In {\em {ICLR}}, 2015.

\bibitem{chen2017rethinking}
Liang-Chieh Chen, George Papandreou, Florian Schroff, and Hartwig Adam.
\newblock Rethinking atrous convolution for semantic image segmentation.
\newblock {\em arXiv preprint arXiv:1706.05587}, 2017.

\bibitem{chen20182}
Yunpeng Chen, Yannis Kalantidis, Jianshu Li, Shuicheng Yan, and Jiashi Feng.
\newblock A\^{} 2-nets: Double attention networks.
\newblock In {\em Advances in neural information processing systems}, pages
  352--361, 2018.

\bibitem{cordts2016cityscapes}
Marius Cordts, Mohamed Omran, Sebastian Ramos, Timo Rehfeld, Markus Enzweiler,
  Rodrigo Benenson, Uwe Franke, Stefan Roth, and Bernt Schiele.
\newblock The cityscapes dataset for semantic urban scene understanding.
\newblock In {\em {CVPR}}, 2016.

\bibitem{everingham2015pascal}
Mark Everingham, SM~Ali Eslami, Luc Van~Gool, Christopher~KI Williams, John
  Winn, and Andrew Zisserman.
\newblock The pascal visual object classes challenge: A retrospective.
\newblock {\em {IJCV}}, 2015.

\bibitem{fu2019dual}
Jun Fu, Jing Liu, Haijie Tian, Yong Li, Yongjun Bao, Zhiwei Fang, and Hanqing
  Lu.
\newblock Dual attention network for scene segmentation.
\newblock In {\em {CVPR}}, pages 3146--3154, 2019.

\bibitem{gao2019global}
Zilin Gao, Jiangtao Xie, Qilong Wang, and Peihua Li.
\newblock Global second-order pooling convolutional networks.
\newblock In {\em Proceedings of the IEEE Conference on Computer Vision and
  Pattern Recognition}, pages 3024--3033, 2019.

\bibitem{hariharan2011semantic}
Bharath Hariharan, Pablo Arbel{\'a}ez, Lubomir Bourdev, Subhransu Maji, and
  Jitendra Malik.
\newblock Semantic contours from inverse detectors.
\newblock In {\em {ICCV}}, 2011.

\bibitem{He2016}
Kaiming He, Xiangyu Zhang, Shaoqing Ren, and Jian Sun.
\newblock Deep residual learning for image recognition.
\newblock In {\em {CVPR}}, 2016.

\bibitem{hou2020strip}
Qibin Hou, Li Zhang, Ming-Ming Cheng, and Jiashi Feng.
\newblock Strip pooling: Rethinking spatial pooling for scene parsing.
\newblock In {\em Proceedings of the IEEE/CVF Conference on Computer Vision and
  Pattern Recognition}, pages 4003--4012, 2020.

\bibitem{howard2019searching}
Andrew Howard, Mark Sandler, Grace Chu, Liang-Chieh Chen, Bo Chen, Mingxing
  Tan, Weijun Wang, Yukun Zhu, Ruoming Pang, Vijay Vasudevan, et~al.
\newblock Searching for mobilenetv3.
\newblock In {\em {ICCV}}, pages 1314--1324, 2019.

\bibitem{howard2017mobilenets}
Andrew~G Howard, Menglong Zhu, Bo Chen, Dmitry Kalenichenko, Weijun Wang,
  Tobias Weyand, Marco Andreetto, and Hartwig Adam.
\newblock Mobilenets: Efficient convolutional neural networks for mobile vision
  applications.
\newblock {\em arXiv preprint arXiv:1704.04861}, 2017.

\bibitem{hu2018gather}
Jie Hu, Li Shen, Samuel Albanie, Gang Sun, and Andrea Vedaldi.
\newblock Gather-excite: Exploiting feature context in convolutional neural
  networks.
\newblock In {\em Advances in neural information processing systems}, pages
  9401--9411, 2018.

\bibitem{hu2018squeeze}
Jie Hu, Li Shen, and Gang Sun.
\newblock Squeeze-and-excitation networks.
\newblock In {\em {CVPR}}, pages 7132--7141, 2018.

\bibitem{huang2018ccnet}
Zilong Huang, Xinggang Wang, Lichao Huang, Chang Huang, Yunchao Wei, and Wenyu
  Liu.
\newblock Ccnet: Criss-cross attention for semantic segmentation.
\newblock {\em arXiv preprint arXiv:1811.11721}, 2018.

\bibitem{li2019hbonet}
Duo Li, Aojun Zhou, and Anbang Yao.
\newblock Hbonet: Harmonious bottleneck on two orthogonal dimensions.
\newblock In {\em Proceedings of the IEEE International Conference on Computer
  Vision}, pages 3316--3325, 2019.

\bibitem{lin2014microsoft}
Tsung-Yi Lin, Michael Maire, Serge Belongie, James Hays, Pietro Perona, Deva
  Ramanan, Piotr Doll{\'a}r, and C~Lawrence Zitnick.
\newblock Microsoft coco: Common objects in context.
\newblock In {\em {ECCV}}, 2014.

\bibitem{linsley2018global}
Drew Linsley, Dan Shiebler, Sven Eberhardt, and Thomas Serre.
\newblock Learning what and where to attend.
\newblock In {\em {ICLR}}, 2019.

\bibitem{liu2018progressive}
Chenxi Liu, Barret Zoph, Maxim Neumann, Jonathon Shlens, Wei Hua, Li-Jia Li, Li
  Fei-Fei, Alan Yuille, Jonathan Huang, and Kevin Murphy.
\newblock Progressive neural architecture search.
\newblock In {\em Proceedings of the European Conference on Computer Vision
  (ECCV)}, pages 19--34, 2018.

\bibitem{liu2018darts}
Hanxiao Liu, Karen Simonyan, and Yiming Yang.
\newblock Darts: Differentiable architecture search.
\newblock {\em arXiv preprint arXiv:1806.09055}, 2018.

\bibitem{liu2020improving}
Jiang-Jiang Liu, Qibin Hou, Ming-Ming Cheng, Changhu Wang, and Jiashi Feng.
\newblock Improving convolutional networks with self-calibrated convolutions.
\newblock In {\em Proceedings of the IEEE/CVF Conference on Computer Vision and
  Pattern Recognition}, pages 10096--10105, 2020.

\bibitem{liu2016ssd}
Wei Liu, Dragomir Anguelov, Dumitru Erhan, Christian Szegedy, Scott Reed,
  Cheng-Yang Fu, and Alexander~C Berg.
\newblock Ssd: Single shot multibox detector.
\newblock In {\em {ECCV}}, 2016.

\bibitem{ma2018shufflenet}
Ningning Ma, Xiangyu Zhang, Hai-Tao Zheng, and Jian Sun.
\newblock Shufflenet v2: Practical guidelines for efficient cnn architecture
  design.
\newblock In {\em Proceedings of the European conference on computer vision
  (ECCV)}, pages 116--131, 2018.

\bibitem{misra2021rotate}
Diganta Misra, Trikay Nalamada, Ajay~Uppili Arasanipalai, and Qibin Hou.
\newblock Rotate to attend: Convolutional triplet attention module.
\newblock In {\em Proceedings of the IEEE/CVF Winter Conference on Applications
  of Computer Vision}, pages 3139--3148, 2021.

\bibitem{mnih2014recurrent}
Volodymyr Mnih, Nicolas Heess, Alex Graves, et~al.
\newblock Recurrent models of visual attention.
\newblock In {\em Advances in neural information processing systems}, pages
  2204--2212, 2014.

\bibitem{park2018bam}
Jongchan Park, Sanghyun Woo, Joon-Young Lee, and In~So Kweon.
\newblock Bam: Bottleneck attention module.
\newblock {\em arXiv preprint arXiv:1807.06514}, 2018.

\bibitem{paszke2019pytorch}
Adam Paszke, Sam Gross, Francisco Massa, Adam Lerer, James Bradbury, Gregory
  Chanan, Trevor Killeen, Zeming Lin, Natalia Gimelshein, Luca Antiga, et~al.
\newblock Pytorch: An imperative style, high-performance deep learning library.
\newblock In {\em {NeurIPS}}, pages 8024--8035, 2019.

\bibitem{real2019regularized}
Esteban Real, Alok Aggarwal, Yanping Huang, and Quoc~V Le.
\newblock Regularized evolution for image classifier architecture search.
\newblock In {\em Proceedings of the aaai conference on artificial
  intelligence}, volume~33, pages 4780--4789, 2019.

\bibitem{russakovsky2015imagenet}
Olga Russakovsky, Jia Deng, Hao Su, Jonathan Krause, Sanjeev Satheesh, Sean Ma,
  Zhiheng Huang, Andrej Karpathy, Aditya Khosla, Michael Bernstein, et~al.
\newblock Imagenet large scale visual recognition challenge.
\newblock {\em {IJCV}}, 115(3):211--252, 2015.

\bibitem{sandler2018mobilenetv2}
Mark Sandler, Andrew Howard, Menglong Zhu, Andrey Zhmoginov, and Liang-Chieh
  Chen.
\newblock Mobilenetv2: Inverted residuals and linear bottlenecks.
\newblock In {\em {CVPR}}, pages 4510--4520, 2018.

\bibitem{selvaraju2017grad}
Ramprasaath~R Selvaraju, Michael Cogswell, Abhishek Das, Ramakrishna Vedantam,
  Devi Parikh, and Dhruv Batra.
\newblock Grad-cam: Visual explanations from deep networks via gradient-based
  localization.
\newblock In {\em {ICCV}}, pages 618--626, 2017.

\bibitem{simonyan2014very}
Karen Simonyan and Andrew Zisserman.
\newblock Very deep convolutional networks for large-scale image recognition.
\newblock In {\em {ICLR}}, 2015.

\bibitem{tan2019mnasnet}
Mingxing Tan, Bo Chen, Ruoming Pang, Vijay Vasudevan, Mark Sandler, Andrew
  Howard, and Quoc~V Le.
\newblock Mnasnet: Platform-aware neural architecture search for mobile.
\newblock In {\em Proceedings of the IEEE Conference on Computer Vision and
  Pattern Recognition}, pages 2820--2828, 2019.

\bibitem{tan2019efficientnet}
Mingxing Tan and Quoc~V Le.
\newblock Efficientnet: Rethinking model scaling for convolutional neural
  networks.
\newblock In {\em ICML}, 2019.

\bibitem{tan2019mixconv}
Mingxing Tan and Quoc~V Le.
\newblock Mixconv: Mixed depthwise convolutional kernels.
\newblock {\em CoRR, abs/1907.09595}, 2019.

\bibitem{tsotsos2011computational}
John~K Tsotsos.
\newblock {\em A computational perspective on visual attention}.
\newblock MIT Press, 2011.

\bibitem{tsotsos1990analyzing}
John~K Tsotsos et~al.
\newblock Analyzing vision at the complexity level.
\newblock {\em Behavioral and brain sciences}, 13(3):423--469, 1990.

\bibitem{wang2020axial}
Huiyu Wang, Yukun Zhu, Bradley Green, Hartwig Adam, Alan Yuille, and
  Liang-Chieh Chen.
\newblock Axial-deeplab: Stand-alone axial-attention for panoptic segmentation.
\newblock {\em arXiv preprint arXiv:2003.07853}, 2020.

\bibitem{wang2018non}
Xiaolong Wang, Ross Girshick, Abhinav Gupta, and Kaiming He.
\newblock Non-local neural networks.
\newblock In {\em {CVPR}}, 2018.

\bibitem{woo2018cbam}
Sanghyun Woo, Jongchan Park, Joon-Young Lee, and In So~Kweon.
\newblock Cbam: Convolutional block attention module.
\newblock In {\em Proceedings of the European conference on computer vision
  (ECCV)}, pages 3--19, 2018.

\bibitem{wu2019fbnet}
Bichen Wu, Xiaoliang Dai, Peizhao Zhang, Yanghan Wang, Fei Sun, Yiming Wu,
  Yuandong Tian, Peter Vajda, Yangqing Jia, and Kurt Keutzer.
\newblock Fbnet: Hardware-aware efficient convnet design via differentiable
  neural architecture search.
\newblock In {\em Proceedings of the IEEE Conference on Computer Vision and
  Pattern Recognition}, pages 10734--10742, 2019.

\bibitem{xie2017aggregated}
Saining Xie, Ross Girshick, Piotr Doll{\'a}r, Zhuowen Tu, and Kaiming He.
\newblock Aggregated residual transformations for deep neural networks.
\newblock In {\em {CVPR}}, 2017.

\bibitem{xu2015show}
Kelvin Xu, Jimmy Ba, Ryan Kiros, Kyunghyun Cho, Aaron Courville, Ruslan
  Salakhudinov, Rich Zemel, and Yoshua Bengio.
\newblock Show, attend and tell: Neural image caption generation with visual
  attention.
\newblock In {\em ICML}, pages 2048--2057, 2015.

\bibitem{zhao2016pyramid}
Hengshuang Zhao, Jianping Shi, Xiaojuan Qi, Xiaogang Wang, and Jiaya Jia.
\newblock Pyramid scene parsing network.
\newblock In {\em {CVPR}}, 2017.

\bibitem{daquan2020rethinking}
Daquan Zhou, Qibin Hou, Yunpeng Chen, Jiashi Feng, and Shuicheng Yan.
\newblock Rethinking bottleneck structure for efficient mobile network design.
\newblock In {\em {ECCV}}, 2020.

\bibitem{zoph2018learning}
Barret Zoph, Vijay Vasudevan, Jonathon Shlens, and Quoc~V Le.
\newblock Learning transferable architectures for scalable image recognition.
\newblock In {\em {CVPR}}, pages 8697--8710, 2018.

\end{thebibliography}
}

\end{document}